\documentclass[conference]{IEEEtran}

\IEEEoverridecommandlockouts
\usepackage{amsmath,amssymb}
\usepackage{booktabs}
\usepackage{array}
\usepackage{graphicx}
\usepackage{xspace}
\usepackage{xcolor}
\usepackage{colortbl}
\usepackage{pifont}
\usepackage[percent]{overpic}

\newcommand{\makecell}[2][c]{\begin{tabular}{@{}#1@{}}#2\end{tabular}}
\newcommand{\method}{OCP-CT\xspace}

\newcommand{\radchest}{RAD-ChestCT\xspace}
\newcommand{\auc}{AUROC\xspace}

\newcommand{\parsection}[1]{\vspace{4pt}\noindent\textbf{#1:}}

\DeclareMathOperator{\softmax}{softmax}

\usepackage[numbers,sort&compress]{natbib}

\renewcommand{\footnoterule}{%
  \kern -3pt
  \hrule width 0.35\columnwidth height 0.4pt
  \kern 2.6pt
}

\definecolor{ieeeblue}{rgb}{0.00,0.30,0.60}
\usepackage[breaklinks,colorlinks,allcolors=ieeeblue]{hyperref}
\usepackage{float}

\title{
Fine-Grained Vision-Language Pretraining with \\ Organ-Conditioned Pattern Tokens for\\  CT Understanding
}

\author{
\IEEEauthorblockN{
Guoliang You\textsuperscript{1},
Xiaomeng Chu\textsuperscript{2,*}
}
\IEEEauthorblockA{
\textsuperscript{1}University of Pennsylvania
}
\IEEEauthorblockA{
\textsuperscript{2}Yale University
}
}

\begin{document}
\maketitle

\begingroup
\renewcommand{\thefootnote}{*}
\footnotetext{Corresponding author.}
\endgroup

\begin{abstract}
Computed tomography (CT) vision-language pretraining from paired volumes and radiology reports is a scalable yet challenging task. Existing methods commonly adopt global scan-report contrast, which is scalable but obscures heterogeneous organ evidence. Meanwhile, direct organ-level alignment remains coarse, since the same anatomy can exhibit multiple distinct radiological appearances. Therefore, pretraining requires a finer alignment unit: the organ-conditioned radiological pattern.
In this work, we propose \method, an organ-conditioned pattern-token alignment framework for CT vision-language pretraining. Specifically, \method preserves a stable global CT-report contrastive branch and introduces an organ pattern interface: sparse Mixture-of-Experts (MoE) routes image and text tokens according to latent radiological patterns, learnable slots query the routed tokens into continuous pattern tokens, and paired token contrast aligns image-text pattern tokens with structured soft targets built from report-derived clinical similarity. On the publicly available CT-RATE and RAD-ChestCT benchmarks, \method achieves average AUROCs of 84.5\% and 69.9\% for zero-shot abnormality diagnosis, respectively. Compared with the strongest prior reported results, these results yield absolute AUROC gains of 6.7 and 0.8 percentage points.
\end{abstract}

\section{Introduction}
\label{sec:intro}

\begin{figure}[t!]
\centering
\includegraphics[width=\linewidth]{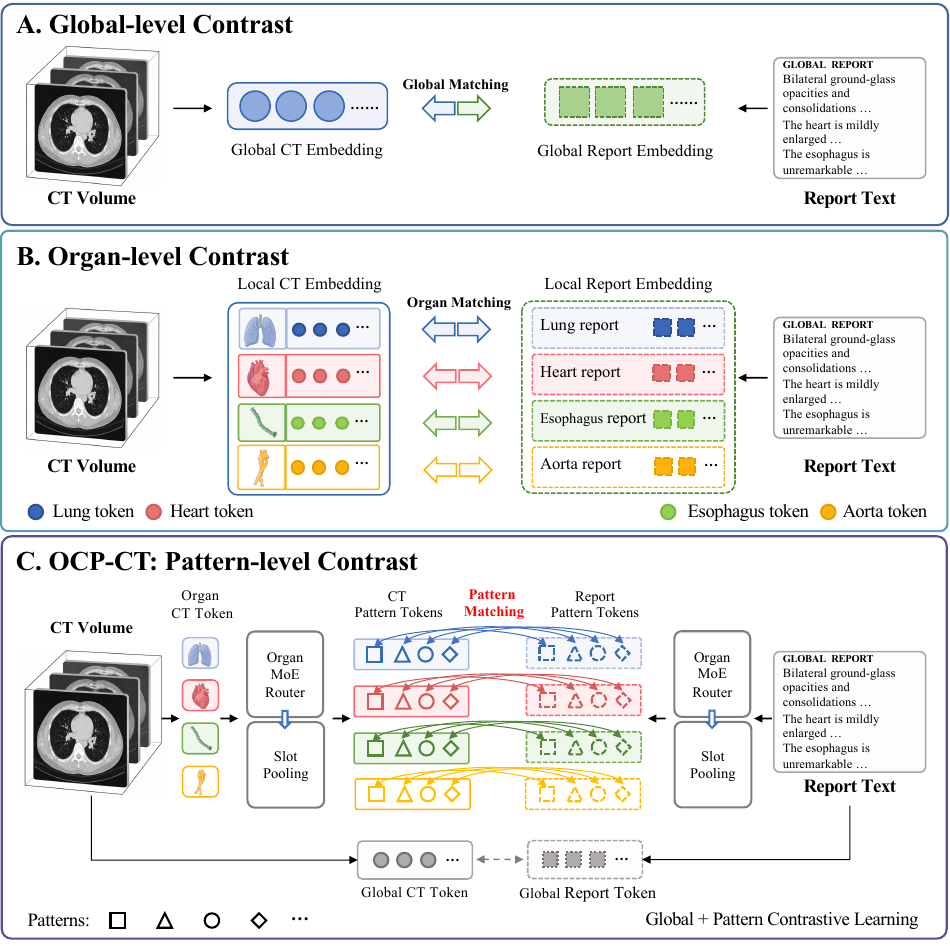}
\caption{\textbf{Motivation of organ-conditioned pattern-token alignment.}
Global-level CT-report contrast compresses organ evidence into one scan-report match, while organ-level alignment can still mix multiple radiological patterns inside the same anatomy.
\method introduces a pattern token bottleneck that routes organ image/text evidence through sparse MoE, pools the routed evidence into compact pattern tokens with organ slots, and aligns paired pattern tokens before the global objective.
}
\label{fig:teaser}
\end{figure}

Computed tomography (CT) is a cornerstone of medical imaging and clinical diagnosis, but interpreting a volumetric study requires reviewing hundreds of slices and reasoning across multiple anatomies~\citep{langlotz2024merlin}. This motivates general-purpose CT representation learning for diagnosis, retrieval, and downstream adaptation. Since dense 3D annotations are costly, recent CT vision-language pretraining (VLP) methods learn scalable representations from routinely collected CT-report pairs~\citep{hamamci2025ctrate,langlotz2024merlin,lai2025brgsa,wald2025colipri}.

A dominant recipe is global-level scan-report contrast, where one CT embedding is aligned with one report embedding. This design is a strong foundation: it inherits the scalability of CLIP-style image-text learning~\citep{radford2021clip} and adapts naturally to radiology reports as weak language supervision~\citep{zhang2020convirt}. Existing 3D CT VLP systems further show that paired CT volumes and reports can support whole-volume diagnosis, retrieval, and general representation learning~\citep{langlotz2024merlin,wald2025colipri}. However, global contrast is coarse: a report may describe findings across lung, pleura, heart, esophagus, and aorta, while the model only receives one scan-report matching signal.
This coarse global supervision can obscure the organ evidence that supports the report. Radiologists typically inspect relevant anatomy, identify local findings, and integrate them into a global-level impression. 

A natural response is to introduce organ-level local alignment in CT vision-language pretraining. Global-local representation methods align image regions with report semantics~\citep{huang2021gloria,muller2022lovt}, and CT-specific methods use fine-grained, multi-grained, or grounded supervision to connect reports with anatomical regions~\citep{shui2025fvlm,chen2024mg3d,zhang2025ctglip,xie2025radgenome}. 
Yet organ-level alignment remains coarse because organ-level features can still mix heterogeneous radiological appearances: the same lung may contain normal parenchyma, diffuse opacity, focal nodules, atelectatic changes, and pleural fluid. The missing unit is therefore the organ-conditioned radiological pattern, not the organ alone. Moreover, local contrast with only paired positives can over-penalize samples sharing related patterns, reflecting the sensitivity of contrastive objectives to positive and negative definitions~\citep{khosla2020supcon,robinson2021hardnegative,shui2025fvlm}. As illustrated in Figure~\ref{fig:teaser}, the key gap is to expose heterogeneous organ radiological patterns while preserving the global CT-report objective.

This motivates a focused question: can CT-report pretraining learn organ radiological patterns while preserving the global volume-report alignment that makes contrastive learning effective? A useful solution should keep the whole-volume contrastive path, decompose organ evidence into compact pattern tokens, and avoid over-penalizing cases that share related radiological patterns.

To answer this question, we propose \method, an organ-conditioned pattern-token framework for CT vision-language pretraining. The central idea is to insert a compact organ-conditioned pattern token bottleneck between dense image/text tokens and the global contrastive objective. Private sparse MoE modules route organ image tokens and organ text evidence through pattern-specialized transformations, and organ slot queries then pool the routed tokens into a fixed number of continuous pattern tokens. Thus, sparse MoE performs pattern triage, while the organ slot queries provide the aggregation interface that yields cross-modal pattern tokens.
The pattern-token interface is trained with paired image--text token contrast, aligning image pattern token $k$ with text pattern token $k$ within the same organ. To avoid forcing all non-paired organ instances into uniformly hard negatives, structured soft targets assign weak positive mass to samples with related patterns.

We evaluate \method on two public CT benchmarks for zero-shot abnormality diagnosis and retrieval. 
The results show improved global CT understanding, while organ-group analysis, module ablations, and patch-occlusion visualization further support the effectiveness of organ-conditioned pattern modeling.

In summary, our main contributions are as follows:
\begin{itemize}
    \item We formulate organ-conditioned pattern-token alignment for CT-report contrastive learning, preserving scalable global volume-report supervision while exposing organ patterns.
    \item We introduce \method, a pattern-token framework that uses sparse MoE and organ slots to route image/text evidence and compress it into paired continuous pattern tokens.
    \item We mitigate hard-negative bias in local contrast with structured soft targets, so samples with related radiological patterns are not treated as uniformly negative.
    \item We demonstrate the advantages of \method through public benchmark evaluation, organ-group analysis, focused ablations, and patch-occlusion visualization.
\end{itemize}

\begin{figure*}[t!]
\centering
\includegraphics[width=\linewidth]{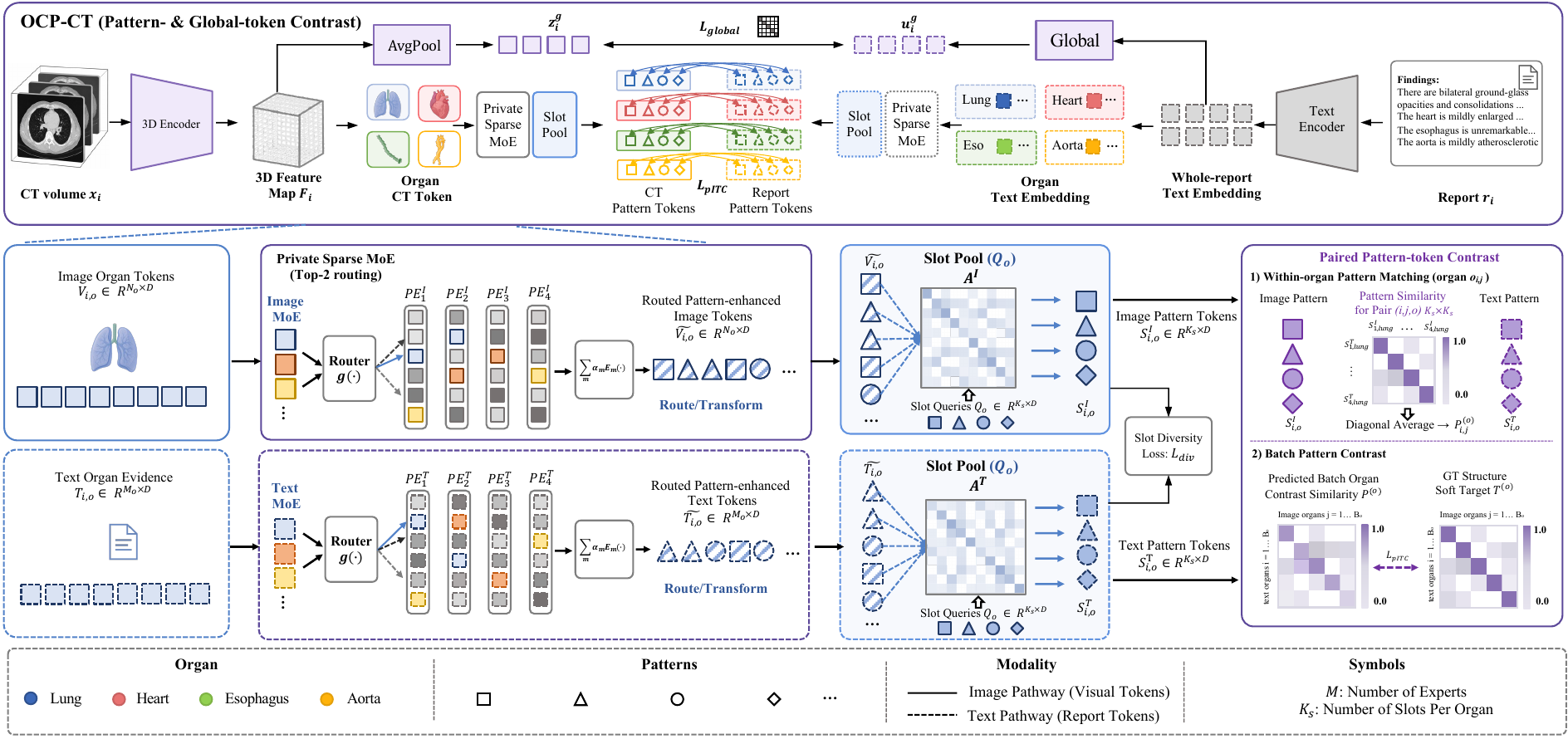}
\caption{\textbf{Overview of \method.} A global branch aligns the CT global token $z_i^g$ with the whole-report token $u_i^g$, while an organ branch selects image organ tokens $V_{i,o}$ and text organ evidence $T_{i,o}$. Private sparse MoE routers send these tokens to $M$ pattern experts, and organ slot queries $Q_o$ pool routed tokens into $K_s$ paired CT/report pattern tokens. Pattern matching and structured soft batch targets define the pattern-level contrast, allowing local evidence to shape the same global representation evaluated downstream. }
\label{fig:method}
\end{figure*}

\section{Related Work}
\label{sec:related}

\paragraph{Global CT Vision-language Pretraining.}
Contrastive vision-language learning has become a standard recipe for learning transferable visual representations from paired text supervision~\citep{radford2021clip,jia2021align,li2021albef}. In radiology, reports provide scalable supervision beyond discrete disease labels, enabling medical image-text pretraining for recognition and retrieval~\citep{zhang2020convirt,sun2022medclip,boecking2022making}. Recent CT systems extend this paradigm to volumetric data, showing that paired CT volumes and reports can support zero-shot diagnosis, retrieval, and generalist representation learning~\citep{hamamci2025ctrate,langlotz2024merlin,lai2025brgsa,wald2025colipri}. These works make global scan-report contrast a strong foundation rather than a weak baseline.

\paragraph{Local and Organ-aware CT Vision-language Pretraining.}
Global-local radiology methods align local visual regions with report semantics to improve label-efficient recognition and localized evaluation~\citep{huang2021gloria,muller2022lovt}. For CT, recent work introduces anatomy, region, or report-structure cues to improve fine-grained understanding and grounding~\citep{shui2025fvlm,chen2024mg3d,zhang2025ctglip,xie2025radgenome}. These studies show that anatomy is a useful handle for CT-report learning. Yet an organ remains coarse: lung evidence may include normal parenchyma, diffuse opacity, nodules, or pleural findings, while report phrases often mix concepts, locations, and attributes. The remaining question is what unit should represent evidence inside each organ.

\paragraph{Sparse Mixture-of-Experts.}
Mixture-of-experts models increase capacity by routing each input to a small subset of expert subnetworks~\citep{shazeer2017moe}. Switch Transformer simplifies sparse expert selection and improves the practicality of conditional computation at large scale~\citep{fedus2022switch}. Beyond language models, V-MoE extends sparse routing to Vision Transformers, showing that image tokens can be adaptively processed by different experts~\citep{riquelme2021vmoe}. Soft MoE further connects expert routing with token mixture and slot-like aggregation by passing weighted combinations of input tokens to experts~\citep{puigcerver2023softmoe}. These works motivate sparse experts as adaptive token processors; in our setting, we use them as organ-conditioned pattern transformation paths for heterogeneous CT image tokens and report evidence.

\section{Method}
\label{sec:method}

\subsection{Overall Framework}

\method learns CT-report representations with organ-conditioned pattern-token alignment. Each example contains a CT volume $x_i$, a paired report $r_i$, multi-abnormality labels $y_i\in\{0,1\}^{18}$, and organ masks when available. The active organ set is
\begin{equation}
\small
\mathcal{O}=\{\text{lung},\text{heart},\text{esophagus},\text{aorta}\}.
\end{equation}
The model keeps a global CT-report contrastive pathway, but augments it with an organ pattern pathway before the final study-level representation is evaluated. The pathway is built around three operations: organ-conditioned token selection, pattern-specialized sparse routing, and continuous pattern-token pooling. The global branch provides the study-level representation used by the main evaluation, while the organ branch shapes this representation through local cross-modal supervision.

Figure~\ref{fig:method} summarizes the data flow. The image encoder (ResNet-18\cite{resnet}) produces both a global token $z_i^g$ and a dense 3D feature map $F_i$. Organ masks select organ image tokens $V_{i,o}$ from the feature map. The text encoder (BioMedVLP CXR-BERT~\cite{boecking2022making}) produces the whole-report token $u_i^g$ and report tokens, from which organ-specific textual evidence $T_{i,o}$ is collected using the report-derived organ fields. Image and text organ tokens are processed by private pattern routers and then compressed independently by parallel Slot Pool blocks that share the organ slot queries $Q_o$. The resulting pattern tokens define within-organ pattern matching and batch pattern contrast shown in the figure; auxiliary prompt and organ-pathology supervision are added in the final training objective.

\subsection{Stable Global Contrastive Backbone}

The global image branch is anchored by average pooling over a 3D CT feature map. Let $F_i\in\mathbb{R}^{C\times D\times H\times W}$ be the layer-4 feature map from the visual encoder. The default whole-volume feature is
\begin{equation}
\small
    a_i=\mathrm{AvgPool}(F_i).
\end{equation}
To allow adaptive readout without replacing the pooled representation, we add a small gated query residual:
\begin{equation}
\small
    z_i^g=W_o\left(\mathrm{AvgPool}(F_i)+\sigma(\gamma)W_q\bar q_i\right),
    \label{eq:qgate}
\end{equation}
where four learnable queries attend to the flattened feature grid, $\bar q_i$ is the mean decoded query, and $\gamma$ is initialized to $-4$. This makes the initial residual scale $\sigma(-4)\approx0.018$. The readout therefore preserves the supervised average-pooled CT representation at the start of contrastive training, while allowing a bounded query residual to adapt the global evidence readout under image-text supervision.

Given the paired text embedding $u_i^g$ from the report encoder, the global contrastive loss follows the standard symmetric form,
\begin{equation}
\footnotesize
\mathcal{L}_{global}=
\frac{1}{2}\left[
\mathrm{CE}\left(\frac{Z^g(U^g)^\top}{\tau_g}, I\right)
+\mathrm{CE}\left(\frac{U^g(Z^g)^\top}{\tau_g}, I\right)
\right],
\end{equation}
where rows are normalized embeddings, $I$ is the paired identity target, and $\tau_g$ is the temperature. All local modules described below are optimized to improve this global representation rather than define a separate test-time model.

\subsection{Organ-Conditioned Pattern Routing}

For each organ $o$, its mask is downsampled to the spatial size of $F_i$, and foreground locations select organ image tokens $V_{i,o}=\{v_{i,o,n}\}_{n=1}^{N_{i,o}}$. Empty masks are skipped. The report is also decomposed into organ-specific text evidence, producing text tokens $T_{i,o}=\{t_{i,o,m}\}_{m=1}^{M_{i,o}}$ from the biomedical text encoder. This gives both modalities the same anatomical conditioning variable, but it does not yet solve the pattern heterogeneity inside an organ.

Because a single organ can contain heterogeneous patterns, \method routes both image and text organ tokens through private sparse pattern experts. Let $PE_m^I$ and $PE_m^T$ denote the $m$-th image and text pattern experts. For an image token $v$, the router computes
\begin{equation}
\small
    p^I(v)=\softmax(W_r^I v), \quad
    \mathcal{K}^I(v)=\mathrm{TopK}(p^I(v),k),
\end{equation}
and the routed token is
\begin{equation}
\small
    \tilde v=\sum_{m\in\mathcal{K}^I(v)}
    \frac{p_m^I(v)}{\sum_{\ell\in\mathcal{K}^I(v)}p_\ell^I(v)}
    PE_m^I(v).
\end{equation}
The text branch uses the same token-wise form with a separate router and expert bank. Applying the routing to all organ tokens gives
\begin{equation}
\small
\tilde V_{i,o}=\{\tilde v_{i,o,n}\}_{n=1}^{N_{i,o}},
\quad
\tilde T_{i,o}=\{\tilde t_{i,o,m}\}_{m=1}^{M_{i,o}}.
\end{equation}
where each routed token is computed by the top-$k$ expert mixture above. In our implementation, $M=4$ experts with top-2 routing provide this token-level processing capacity.
We view the expert bank as learned pattern-specialized transformation paths: organ masks provide anatomical conditioning, while the router selects latent token processors rather than predefined organ or disease classes. A load-balancing auxiliary loss discourages collapse to a single path.

This routing stage addresses heterogeneity at the token-transformation level. Each organ token receives a small set of learned processing paths before any fixed-size local representation is formed. The next step is therefore aggregation, which converts the routed pattern-enhanced organ tokens into the pattern tokens used for cross-modal alignment.

\subsection{Continuous Pattern Tokens}

After sparse routing, learnable slot queries compress the routed pattern-enhanced organ tokens into a fixed set of continuous pattern tokens. In contrast to MoE, which selects transformation paths, Slot Pool performs aggregation: each slot query attends to the routed token set and summarizes recurring evidence into one latent component. Let $Q_o\in\mathbb{R}^{K_s\times C}$ be the slot queries for organ $o$. Slot Pool first forms an attention matrix from slot queries to routed tokens:
\begin{equation}
\small
A^I_{i,o}
=
\softmax\left(
\frac{(Q_o W_q^I)(\tilde V_{i,o}W_k^I)^\top}{\sqrt{d}}
\right).
\end{equation}
The text-side attention $A^T_{i,o}$ has the same form over $\tilde T_{i,o}$. For each organ, Slot Pool produces image pattern tokens and text pattern tokens,
\begin{equation}
\small
S_{i,o}^I=A^I_{i,o}\tilde V_{i,o},\quad
S_{i,o}^T=A^T_{i,o}\tilde T_{i,o},
\end{equation}
where $S_{i,o}^I,S_{i,o}^T\in\mathbb{R}^{K_s\times d}$ and $K_s=6$ in our implementation; $K_s$ sets the capacity of the pattern-token bottleneck for summarizing routed organ evidence.

The pattern tokens are latent components. We do not assign pattern token $k$ a fixed clinical label such as location or severity. Instead, the tokens are shaped by anatomical masks, sparse MoE, slot diversity regularization, organ text supervision, pathology heads, and paired image-text token contrast. Since multiple slots would otherwise be free to copy the same dominant organ feature, we apply a diversity penalty to normalized image slots:
\begin{equation}
\small
    \mathcal{L}_{div}
    =
    \frac{1}{K_s(K_s-1)}
    \sum_{k\neq l}
    \left|
    \left\langle
    \hat S^I_{i,o,k},\hat S^I_{i,o,l}
    \right\rangle
    \right|.
\end{equation}
The penalty does not force human-named semantics, but it encourages the slot interface to carry multiple local components rather than a repeated organ-level summary.

\subsection{Paired Pattern-Token Contrast}

The local cross-modal objective has two parts: within-organ pattern matching and batch pattern contrast.

\paragraph{Within-organ Pattern Matching.}
For image sample $i$ and report sample $j$ in organ $o$, the within-organ pattern matching score is
\begin{equation}
\small
    p_{ij}^{(o)}=\frac{1}{K_s}\sum_{k=1}^{K_s}
    \left\langle
    \frac{S^{I}_{i,o,k}}{\|S^{I}_{i,o,k}\|_2},
    \frac{S^{T}_{j,o,k}}{\|S^{T}_{j,o,k}\|_2}
    \right\rangle .
    \label{eq:patternmatch}
\end{equation}
This diagonal matching turns the pattern-token index into a shared image-text interface. Each image pattern token is trained to match the corresponding text pattern token for the same organ, so the local objective preserves multiple aligned evidence components instead of collapsing the organ into a single pooled descriptor.

\paragraph{Structured Batch Target.}
The scores $p_{ij}^{(o)}$ form the predicted batch pattern-similarity matrix
$P^{(o)}\in\mathbb{R}^{B_o\times B_o}$, where $P_{ij}^{(o)}=p_{ij}^{(o)}$.
For each organ mini-batch, we construct a structured target matrix $T^{(o)}$ that encodes pairwise clinical similarity between samples:
\begin{equation}
\small
    T^{(o)} = \lambda_d I
    + \lambda_a A^{abn}
    + \lambda_c A^{concept}
    + \lambda_l A^{loc}
    + \lambda_r A^{attr}.
    \label{eq:structsoft}
\end{equation}
Here $I$ is the paired identity target. $A^{abn}$ is derived from the 18 abnormality labels, $A^{concept}$ measures overlap between organ-mapped disease labels, and $A^{loc}$ and $A^{attr}$ encode coarse location and attribute agreement extracted from report fields. The report fields are obtained with RadGraph-XL-assisted parsing and LLM normalization~\citep{bui2024radgraphxl,yang2025qwen3}. We use $\lambda_d=1.0$, $\lambda_a=0.8$, $\lambda_c=0.8$, $\lambda_l=0.20$, and $\lambda_r=0.15$.

\paragraph{Batch Pattern Contrast.}
After row normalization, $\bar T^{(o)}$ serves as the soft label distribution for contrastive classification over the predicted matrix $P^{(o)}$. The image-to-text batch pattern contrast is
\begin{equation}
\small
\mathcal{L}_{I\rightarrow T}^{(o)}
=-\frac{1}{B_o}\sum_i\sum_j \bar T_{ij}^{(o)}
\log
\frac{\exp(P_{ij}^{(o)}/\tau_l)}
{\sum_m \exp(P_{im}^{(o)}/\tau_l)}.
\end{equation}
The text-to-image term uses $\bar T^{(o)\top}$, and $\mathcal{L}_{pITC}$ averages the two directions. This objective uses the structured target as a soft label distribution, keeping the paired study dominant while allowing related off-diagonal organ samples to contribute weak supervision.

% EXP
\begin{table*}[!t]
\caption{\textbf{Zero-shot abnormality diagnosis on public CT benchmarks.} Diagnostic performance is reported using AUROC, accuracy, F1, and precision. The best available value in each column is highlighted in bold. COLIPRI-C$^\dagger$ denotes the contrastive-only COLIPRI variant evaluated under the comparable zero-shot diagnosis setting.}
\label{tab:diagnosis_results}
\centering
\footnotesize
\setlength{\tabcolsep}{16pt}
\begin{tabular}{l|cccc|cccc}
\toprule
Method & \multicolumn{4}{c|}{CT-RATE$\uparrow$} &
\multicolumn{4}{c}{RAD-ChestCT$\uparrow$} \\
\cline{2-9}
 & \auc & ACC & F1 & Prec. & \auc & ACC & F1 & Prec. \\
\midrule
CT-Net~\citep{draelos2021radchestct} & 60.3 & 58.1 & 63.1 & 23.9 & 54.4 & 54.0 & 58.7 & 28.5 \\
CT-CLIP~\citep{hamamci2025ctrate} & 73.1 & 66.8 & 70.7 & 32.3 & 62.9 & 59.5 & 64.2 & 33.6 \\
BIUD~\citep{cao2024biud} & 71.3 & 68.1 & 71.6 & 33.8 & 62.9 & 60.6 & 65.2 & 33.7 \\
Merlin~\citep{langlotz2024merlin} & 72.8 & 67.2 & 70.9 & 33.7 & 64.4 & 61.9 & 66.3 & 34.8 \\
COLIPRI-C$^\dagger$~\citep{wald2025colipri} & 76.3 & -- & -- & -- & 69.1 & -- & -- & -- \\
fVLM~\citep{shui2025fvlm} & 77.8 & 71.8 & 75.1 & 37.9 & 68.0 & 64.7 & 68.8 & 37.4 \\
\midrule
\rowcolor[gray]{.92}\textbf{\method} & \textbf{84.5} & \textbf{78.1} & \textbf{80.3} & \textbf{44.5} & \textbf{69.9} & \textbf{65.2} & \textbf{69.7} & \textbf{47.0} \\
\bottomrule
\end{tabular}
\end{table*}
% EXP

\subsection{Pathology Prompt Anchoring}

The prompt anchor is complementary to the pattern-level image--text contrast above. Pattern-level contrast aligns each organ image pattern token with the corresponding organ report pattern token and uses structured targets to relax batch negatives. This provides case-level cross-modal alignment, but its supervision still comes from instance-specific report text.

We add a label-conditioned anchor by encoding positive and negative pathology prompts with the same text-side pattern-token branch. Image pattern tokens and report pattern tokens are compared with these prompt pattern tokens using pathology labels or organ-text parses as binary targets. A shallow organ pathology head also reads the mean image pattern token for organ-mapped findings. These auxiliary terms correspond to $\mathcal{L}_{prompt}$ and $\mathcal{L}_{organCls}$ in the final objective; they keep the pattern-token interface tied to named abnormality cues during training, but are not separate evaluation heads or expert-verified pattern labels.

\subsection{Training Objective}

The method is built on a supervised visual foundation and optimized for CT-report contrastive representation learning. The visual encoder is initialized by supervised multi-label CT abnormality training without the report-contrastive objective, after which the full model is optimized with the losses above. We optimize the following final objective:
\begin{equation}
\small
\begin{aligned}
\mathcal{L}={}&
\mathcal{L}_{global}
+\lambda_{local}\mathcal{L}_{pITC}
+ \lambda_{div}\mathcal{L}_{div}\\
&+ \lambda_{moe}\mathcal{L}_{moe}
+\mathcal{L}_{prompt}
+\mathcal{L}_{organCls}.
\end{aligned}
\end{equation}
$\mathcal{L}_{moe}$ is the router load-balancing loss. $\mathcal{L}_{prompt}$ uses binary cross-entropy to anchor image/report pattern tokens to positive and negative pathology prompts, while $\mathcal{L}_{organCls}$ is a binary multi-label classification loss for organ-mapped findings from pooled pattern tokens.

\section{Experiments}
\label{sec:experiments}

\subsection{Datasets and Metrics}
\parsection{Datasets}
We evaluate on two public chest CT benchmarks. CT-RATE~\citep{hamamci2025ctrate} contains 50,188 chest CT scans from 21,304 patients with 18 abnormality labels. We adopt the CT-CLIP training/validation splits and use CT-RATE for zero-shot abnormality diagnosis and retrieval experiments. For external validation, we use RAD-ChestCT~\citep{draelos2021radchestct}, which comprises 36,316 non-contrast chest CT scans from Duke University collected between 2012 and 2017 with 83 abnormality labels. Following the CT-CLIP evaluation framework, we evaluate on its public subset of 3,630 scans. Since RAD-ChestCT is collected from a different institution and has a different label set, it provides an external check for zero-shot diagnosis.

\parsection{Metrics}
We consider three evaluation tasks: zero-shot abnormality diagnosis, image-image retrieval, and report-image retrieval. These tasks cover both diagnostic prediction and retrieval-based representation evaluation. For diagnosis, we report \auc, ACC, F1, and precision. For image-image retrieval, we report MAP@5/10/50. For report-image retrieval, we report Recall@5/10/50/100. All metrics follow the standard evaluation protocol used by the compared baselines, and missing baseline metrics are marked as ``--''.

\subsection{Main Results}

\parsection{Zero-shot Abnormality Diagnosis}
Table~\ref{tab:diagnosis_results} compares zero-shot abnormality diagnosis on CT-RATE and \radchest. On CT-RATE, \method achieves an \auc of 84.5, with an accuracy, F1-score, and precision of 78.1, 80.3, and 44.5, respectively. On the external \radchest benchmark, \method obtains an \auc of 69.9, with an accuracy, F1-score, and precision of 65.2, 69.7, and 47.0, respectively. These results indicate that the learned representation remains effective for prompt-based abnormality diagnosis across both the in-domain benchmark and external validation.
Compared with fVLM, the most related anatomy-aware CT VLP baseline with complete reported metrics, \method improves CT-RATE by 6.7, 6.3, 5.2, and 6.6 percentage points in \auc, accuracy, F1-score, and precision, respectively, and improves \radchest by 1.9, 0.5, 0.9, and 9.6 percentage points, respectively. The consistent gains over an anatomy-level alignment method suggest that modeling multiple organ-conditioned pattern tokens provides benefits beyond assigning one pooled representation to each anatomy. \method also exceeds COLIPRI-C in the available \auc columns. Since the diagnosis is performed from the final scan-level representation, these improvements show that local pattern supervision contributes to the global embedding used at inference. This supports the design goal of using organ pattern tokens as training-time evidence interfaces.

\begin{table}[!t]
\caption{\textbf{CT-RATE retrieval.} Image-image retrieval measures abnormality-based neighborhood structure, while report-image retrieval measures cross-modal alignment. \method is evaluated under the same retrieval setting.}
\label{tab:retrieval_results}
\centering
\footnotesize
\setlength{\tabcolsep}{4pt}
\begin{tabular}{p{0.25\linewidth}|p{0.32\linewidth}p{0.32\linewidth}}
\toprule
Method &
\makecell{Img2Img MAP\\@5/10/50$\uparrow$} &
\makecell{Rpt2Img Recall\\@5/10/50/100$\uparrow$} \\
\midrule
CT-Net~\citep{draelos2021radchestct} & 59.4 / 48.1 / 40.7 & -- / -- / -- / -- \\
VocabFine~\citep{hamamci2025ctrate} & 68.3 / 57.2 / 48.8 & 0.1 / 0.6 / 2.3 / 2.0 \\
ClassFine~\citep{hamamci2025ctrate} & 67.9 / 56.8 / 48.5 & -- / -- / -- / -- \\
Merlin~\citep{langlotz2024merlin} & 62.6 / 51.3 / 43.9 & 1.5 / 2.7 / 7.7 / 12.7 \\
CT-CLIP~\citep{hamamci2025ctrate} & 68.3 / 57.2 / 48.9 & 2.9 / 5.0 / 18.0 / 28.7 \\
\midrule
\rowcolor[gray]{.92}\textbf{\method} & \textbf{70.7 / 61.0 / 53.5} & \textbf{7.4 / 12.4 / 34.3 / 48.6} \\
\bottomrule
\end{tabular}
\end{table}

\parsection{Image-image Retrieval}
Table~\ref{tab:retrieval_results} evaluates whether the scan-level embedding preserves abnormality-based neighborhood structure. \method obtains MAP@5, MAP@10, and MAP@50 scores of 70.7, 61.0, and 53.5 on CT-RATE, respectively, outperforming the strongest listed prior result by 2.4, 3.8, and 4.6 points, respectively. This shows that the organ-conditioned pattern pathway improves the final global CT embedding, rather than only adding an auxiliary local training signal.
The improvement remains consistent as the retrieval depth increases, with the largest gain observed at MAP@50. Since image-image retrieval ranks CT volumes by embedding similarity and evaluates whether clinically related scans are retrieved near the query, the stronger MAP@50 suggests that \method preserves disease-neighborhood structure over a broader retrieved set.

\parsection{Report-image Retrieval}
Report-image retrieval evaluates cross-modal alignment by using a report as the query and checking whether its paired CT volume appears among the top-ranked image results. 
\method reaches Recall@5, Recall@10, Recall@50, and Recall@100 scores of 7.4, 12.4, 34.3, and 48.6, respectively, outperforming the strongest listed prior result by 4.5, 7.4, 16.3, and 19.9 points, respectively.
COLIPRI-C retrieval is omitted because it was not produced under the same configuration.
These gains indicate that the final scan-level embedding is better aligned with report semantics. Since the retrieval target is the paired volume rather than a disease category, improved report-image recall provides complementary evidence for cross-modal representation learning. This is consistent with the paired pattern-token objective, which aligns organ image pattern tokens with organ report pattern tokens and, through the shared visual backbone, shapes the scan-level representation used for retrieval.

\begin{figure}[b]
\centering
\includegraphics[width=\linewidth]{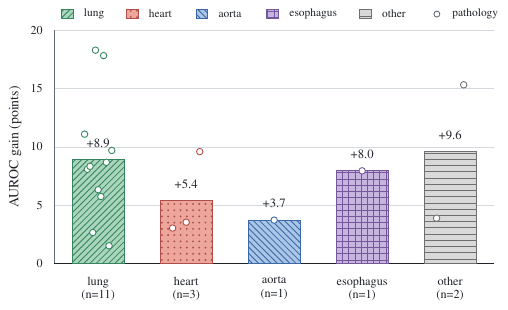}
\caption{\textbf{Organ-level AUROC gains over the global-only baseline.} Bars report mean AUROC improvement in percentage points from the matched global-only baseline to \method within each organ group, and open circles mark the individual abnormality-label gains in that group. $n$ denotes the number of abnormality labels. Other labels are shown separately because they are not assigned to the four active organs.}
\label{fig:organ_auc_gain}
\end{figure}

\parsection{Organ-group Behavior}
Fig.~\ref{fig:organ_auc_gain} summarizes CT-RATE diagnosis gains from the matched global-only baseline to \method across organ groups. Bars show mean AUROC improvement within each group. All analyzed groups show positive mean gains. The other group is reported separately because Medical material and Lymphadenopathy are not assigned to the four active organs.
The mean AUROC gains for lung, heart, aorta, esophagus, and other labels are 8.9, 5.4, 3.7, 8.0, and 9.6 points, respectively.
The label-level points indicate that the improvement is not concentrated in a single abnormality group. Overall, the result suggests that adding organ-conditioned pattern supervision improves the shared scan-level representation across several clinically meaningful label groups, rather than only benefiting one dominant organ category.

\subsection{Qualitative Visualization}

\begin{figure*}[!t]
\centering
\includegraphics[width=\textwidth]{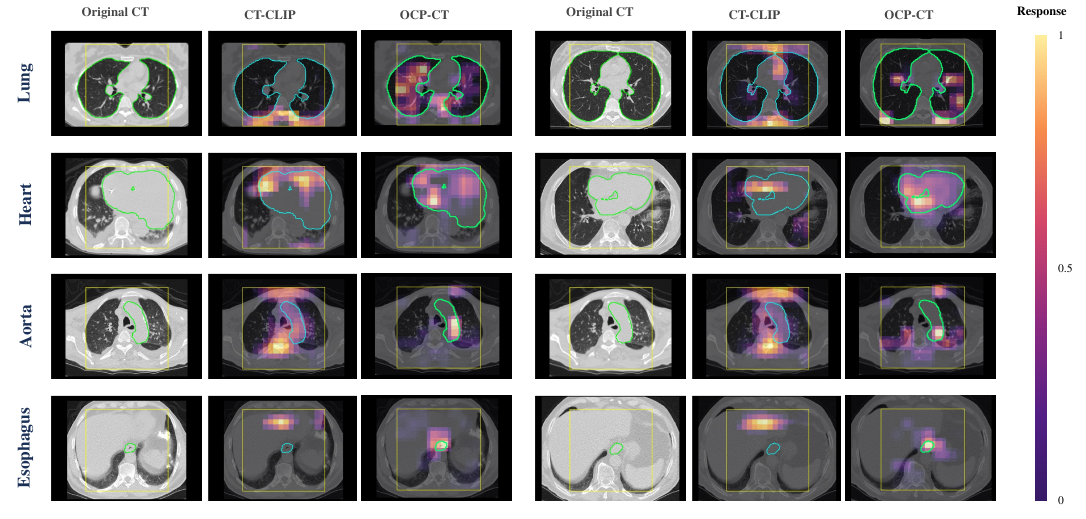}
\caption{\textbf{Patch-occlusion visualization.} Selected cases from each available organ group compare the original CT slice, CT-CLIP, and \method. The colorbar reports normalized occlusion response; warmer colors correspond to larger positive disease-score drops after local patch occlusion. The yellow box marks the model-visible crop, and heatmaps are only defined inside this crop. \method tends to concentrate stronger responses on organ-related regions, suggesting that the prediction is more sensitive to relevant organ evidence.}
\label{fig:occlusion}
\end{figure*}

\parsection{Patch-occlusion Evidence}
Fig.~\ref{fig:occlusion} visualizes patch-occlusion sensitivity for representative cases across the active organ groups. For each case, local patches are occluded and the change in the target disease score is recorded; warmer responses indicate patches whose removal causes a larger score drop. Compared with CT-CLIP, \method shows more concentrated responses within the model-visible crop and closer to the organ regions associated with the target abnormality. The yellow box marks the crop observed by the model, and the heatmaps are interpreted only within this visible region.

This visualization further supports the role of organ-conditioned pattern supervision in shaping the global prediction. Since both methods are evaluated through their global scores, stronger organ-related occlusion responses suggest that the learned representation is more sensitive to organ-related radiological evidence, rather than reflecting only a separate local branch. This model-agnostic sensitivity check complements the quantitative diagnosis, retrieval, and ablation results.

\subsection{Ablation Study}

\begin{table}[!t]
\caption{\textbf{Core \method ablation.} Diagnosis AUROC is reported under the same zero-shot abnormality diagnosis setting as Table~\ref{tab:diagnosis_results}. The full \method row is followed by a global-branch-only variant that removes the organ-conditioned \method pathway, then module-removal variants for sparse routing, pattern-token slots, paired token contrast, and structured soft targets.}
\label{tab:core_ablation}
\centering
\footnotesize
\setlength{\tabcolsep}{15pt}
\begin{tabular}{p{0.6\linewidth}|c}
\toprule
Variant &
\makecell{Diagnosis\\\auc$\uparrow$} \\
\midrule
\rowcolor[gray]{.92}\textbf{\method full} & \textbf{84.49} \\
global branch only (w/o organ \method) & 76.41 \\
w/o sparse pattern routing & 83.35 \\
w/o pattern-token slots & 83.71 \\
w/o paired token contrast & 82.30 \\
w/o structured soft targets & 81.24 \\
\bottomrule
\end{tabular}
\end{table}

Table~\ref{tab:core_ablation} evaluates the main components of \method under the same CT-RATE diagnosis setting. The ablations are designed to separate the effect of the global branch, the organ-conditioned pattern-token bottleneck, and the contrastive target design.

\parsection{Global Branch vs. Organ-conditioned Pathway}
Removing the organ-conditioned pattern-token pathway leaves only the global branch and reduces AUROC from 84.49 to 76.41. This supports the motivation that global scan-report alignment is a strong foundation, and that adding organ pattern supervision substantially improves the final representation. The gap between the global-only variant and full \method indicates that the organ-conditioned pathway provides complementary training signals beyond whole-volume contrast alone.

\parsection{Sparse Routing and Pattern-token Slots}
Removing sparse pattern routing decreases AUROC to 83.35, while removing pattern-token slots gives 83.71. These results indicate that both parts of the pattern-token bottleneck are useful: sparse routing provides input-dependent token transformation, and slot pooling compresses routed organ evidence into a fixed set of pattern tokens. The drop without slots is consistent with our motivation that an organ should not be represented only by one pooled descriptor. Together, these components provide the routing-and-aggregation mechanism needed to organize heterogeneous organ evidence into pattern-level representations.

\parsection{Paired Contrast and Structured Soft Targets}
The larger drops come from removing the contrastive interface or its target structure. Without paired token contrast, AUROC decreases to 82.30; without structured soft targets, it further decreases to 81.24. This suggests that pattern-token contrast works best when fine-grained image--report matching is paired with structured clinical similarity. Paired pattern tokens define the local image--text similarity space, while soft targets allow samples with related abnormalities, concepts, or locations to contribute weak positive supervision instead of being treated as uniformly hard negatives. They provide the pattern-token objective with both fine-grained cross-modal matching and clinically informed supervision.

\section{Discussion and Limitations}
\label{sec:discussion}

\paragraph{What the Results Establish.}
\method is designed to test whether organ-conditioned pattern tokens can improve the global CT-report representation, rather than merely adding an auxiliary local branch. The evaluation therefore uses standard scan-level endpoints: zero-shot abnormality diagnosis, image-image retrieval, and report-image retrieval. Together with the global-only ablation, the results show that local pattern supervision can improve the shared scan-level representation used at inference. The claim is not that the learned tokens are expert-defined radiological categories, but that exposing organ pattern structure during training benefits global CT VLP endpoints.

\paragraph{Limitations.}
The local supervision is automatically derived from reports and labels, so errors in entity extraction, organ assignment, location parsing, or attribute normalization may affect the structured targets. The current organ branch covers lung, heart, esophagus, and aorta, leaving other findings to the global pathway. In addition, external validation is currently limited to abnormality diagnosis on \radchest; retrieval transfer is evaluated only on CT-RATE. Patch-occlusion visualizations provide weakly supervised sensitivity evidence, but do not replace expert localization annotations.

\paragraph{Implication.}
These results suggest a practical middle ground between purely global CT-report contrast and fully grounded supervision. \method keeps the scalable global scan-report objective, while using weak organ structure and pattern tokens to shape the representation through a shared backbone. This design allows organ-local evidence to guide global representation learning without requiring dense expert lesion annotations.

\section{Conclusion}
\label{sec:conclusion}

We presented \method, an organ-conditioned pattern-token alignment framework for CT-report contrastive pretraining. \method preserves global scan-report alignment while introducing an organ-local pattern interface, where a sparse MoE module routes image and text evidence, organ slots form continuous pattern tokens, and structured soft targets supervise paired pattern-token contrast. Across CT-RATE and RAD-ChestCT, \method improves zero-shot abnormality diagnosis, and on CT-RATE it also improves image-image and report-image retrieval using the same final scan-level representation. The ablation and organ-group analyses further show that the gains come from the organ-conditioned pathway and its pattern-level contrastive design, rather than from a separate task-specific local head. Overall, these results suggest that CT VLP benefits from representing organ evidence at the pattern-token level before optimizing the final global scan-report representation. This leads to stronger diagnosis and retrieval performance while preserving the scalable global contrastive training paradigm. 
Future work will explore broader anatomical coverage and richer report-derived attributes to further improve the generality of organ-conditioned CT representation learning across datasets.
{
    \bibliographystyle{IEEEtran}
    \bibliography{main}
}
\end{document}